\documentclass{article}
\usepackage{iclr2027_conference,times}
\iclrfinalcopy
\usepackage{amsmath,amsfonts}

\usepackage{hyperref}
\usepackage{cleveref}
\usepackage{url}
\usepackage{marvosym}
\usepackage{graphicx}
\usepackage[table,dvipsnames]{xcolor}
\usepackage{nicematrix}
\usepackage{makecell}
\usepackage{placeins}
\usepackage{tikz}
\ifdefined\pdfsuppresswarningpagegroup
\fi
\newcommand{\lmttbf}[1]{{\fontfamily{lmtt}\selectfont\bfseries #1}}

\title{Collaborative Learning and Reasoning \\ over Model Slices}

\author{%
\hspace*{-0.75em}Yuheng Wang{$^\dagger$} \quad Runde Yang{$^\dagger$} \hspace{0.5em} Zhuohan Bao \hspace{0.5em} Qiran Zhang \hspace{0.5em} Huatao Li  \\
\hspace*{-0.75em}\textbf{Yiyang Zheng}\quad \textbf{Lin Wu} \quad \textbf{Jingru Fan} \quad \textbf{Shu Yao} \quad \textbf{Yihan Li} \quad \textbf{Zhi Li} \quad \textbf{Chen Qian}{\textsuperscript{\Letter}}
\\
\vspace{-0.5em} \\
\hspace*{-0.75em}\textnormal{School of Artificial Intelligence, Shanghai Jiao Tong University}}

\begin{document}
\AddToShipoutPictureFG*{%
  \AtPageUpperLeft{%
    \put(505,-14){\makebox[0pt][r]{\raisebox{-\height}{\includegraphics[height=0.28in]{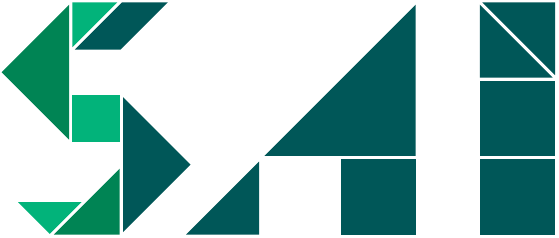}}}}%
  }%
}%
\maketitle
\begingroup
\renewcommand\thefootnote{}
\makeatletter
\def\Hy@footnote@currentHref{author.equal-contribution}
\makeatother
\footnotetext{$^\dagger$ Equal contribution.}
\makeatletter
\def\Hy@footnote@currentHref{author.corresponding}
\makeatother
\footnotetext{\Letter\ Corresponding author.}
\endgroup
\setcounter{footnote}{0}
\lhead{Preprint.}

\begin{abstract}

Large language models (LLMs) excel across diverse tasks, benefiting from knowledge priors, in-context reasoning, and test-time computation. Despite these strengths, single-model reasoning remains vulnerable to error propagation, as later steps depend on the model's earlier outputs, leaving reasoning blind spots difficult to independently reassess. To address this limitation, we develop \textit{collaborative learning}, encompassing the construction of heterogeneous collaborators and the learning of effective cross-model collaboration. Specifically, \textit{model slicing} replaces a large model with smaller, heterogeneous models under a comparable parameter budget, enabling prior reasoning to be reassessed across distinct model parameterizations. \textit{Pyramid supervision} combines protocol distillation with step-level supervision to refine reasoning and interaction decisions throughout collaboration. We evaluate our approach across four task categories: multi-hop question answering, mathematical reasoning, code generation, and open-ended constrained generation. The trained heterogeneous models achieve comparable or better task performance than the original large model and established multi-agent baselines, including reinforcement-learning-based methods, while also achieving higher inference efficiency than the multi-agent baselines and exhibiting differentiated collaboration patterns across tasks. These findings show that learned collaboration turns model heterogeneity into complementary reasoning capabilities. Code is available at: \texttt{\href{https://github.com/SJTUFreshman/collaborative-learning}{github.com/SJTUFreshman/collaborative-learning}}.

\end{abstract}

\section{Introduction}

Large language models (LLMs) have demonstrated capabilities across tasks~\citep{wang2024mmlupro}. Advances highlight the benefits of scaling test-time computation~\citep{snell2025scaling,muennighoff2025s1,wu2025inferencescaling}. Many approaches seek to improve task performance by allowing a single model to structure its reasoning process~\citep{besta2024graphthoughts} and generate, inspect, and revise intermediate steps~\citep{zhou2024selfdiscover,zhang2024restmcts,kumar2025selfcorrect}. However, when the same model both generates and evaluates a solution, it may exhibit self-bias, favoring its own outputs and limiting its ability to identify and correct reasoning errors~\citep{huang2024selfcorrect,xu2024selfbias}. Consequently, errors in intermediate steps may remain undetected and influence subsequent reasoning~\citep{zheng2025processbench}. This motivates \textit{reasoning-state isolation} across models, allowing prior reasoning to be independently reassessed under distinct model parameterizations rather than repeatedly evaluated within the same model's reasoning space~\citep{du2024debate,liang2024mad}.

\begin{figure}[!t]
\centering
\includegraphics[width=\linewidth]{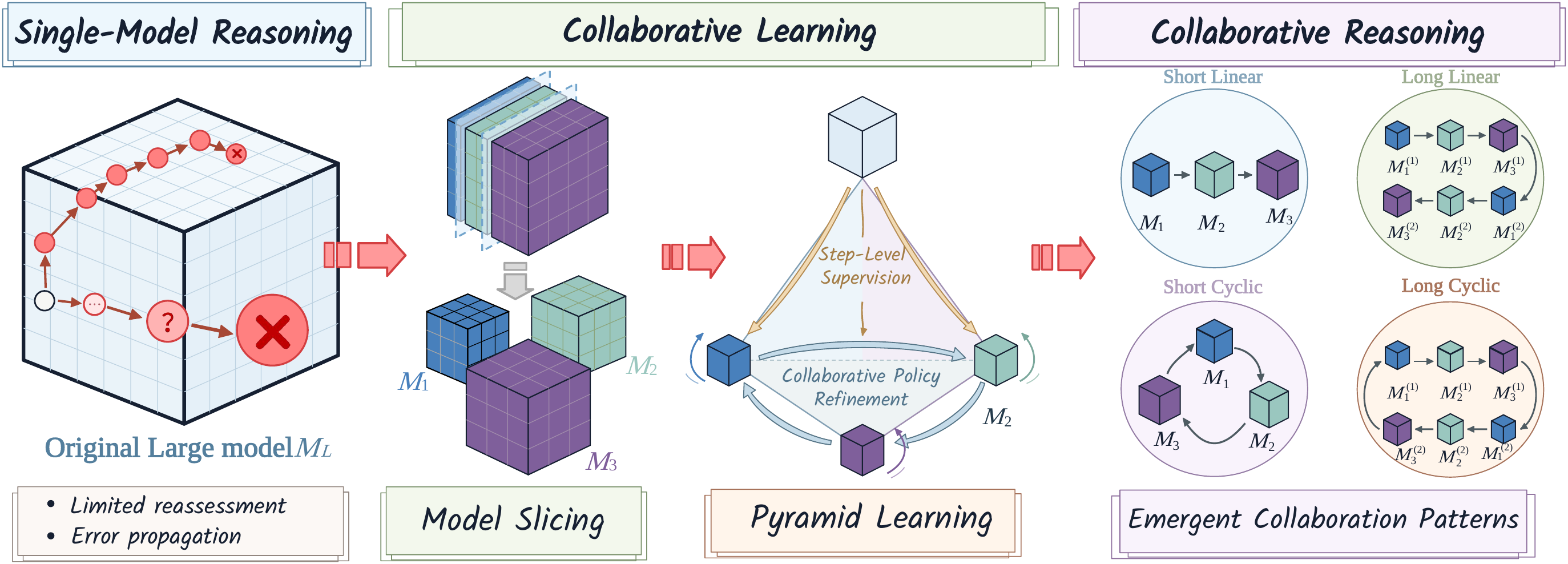}
\vspace{-1.5em}
\caption{\textbf{Collaborative Learning and Reasoning over Model Slices.}
Model slicing reallocates a single large model into multiple smaller, heterogeneous models under a comparable parameter budget, enabling reasoning to be reassessed across model parameterizations. Building on this, collaborative learning trains heterogeneous models through pyramid supervision, combining protocol distillation for interaction initialization with step-level supervision over collaboration trajectories\protect\footnotemark.}
\label{fig:model_slicing_concept}
\vspace{-0.8em}
\end{figure}

Accordingly, we develop a \textit{collaborative learning} framework for reasoning and learning across heterogeneous models, as illustrated in Figure~\ref{fig:model_slicing_concept}. Model slicing replaces a single large model with several smaller, heterogeneous models under a comparable total parameter budget~\citep{liu2026collaboration}. Through explicit handoffs of task state between models~\citep{chen2024agentverse}, prior reasoning can be independently reassessed under distinct model parameterizations, thereby enabling model-specific reasoning-state isolation while producing interaction trajectories that capture reasoning and interaction decisions. Together, this isolation and the resulting trajectories provide the structural basis for training heterogeneous models to leverage complementary capabilities~\citep{wang2025mixture,abdulaal2025balancing,chen2025ioa} when questioning, extending, or revising prior conclusions.

Although this structure provides a foundation for cross-model collaboration, outcome-level supervision alone provides insufficient guidance for intermediate reasoning and interaction decisions throughout the collaborative process~\citep{lightman2024verify,wang2024mathshepherd,setlur2025progress}. To address this limitation, we propose \textit{pyramid supervision}, which uses supervisory signals from larger models to guide the heterogeneous models produced by model slicing toward effective cross-model collaboration. It combines interaction initialization with fine-grained supervision over reasoning and interaction decisions along model-generated collaboration trajectories, while integrating task-level rewards to jointly optimize intermediate decision quality and overall task performance. Through this training mechanism, heterogeneous models learn effective collaborative reasoning and develop differentiated interaction patterns across tasks.

We evaluate our approach across four task categories: multi-hop question answering, mathematical reasoning, code generation, and open-ended constrained generation. With a total parameter budget comparable to that of the original large model, the trained heterogeneous models collectively match or outperform the original model as well as established multi-agent system (MAS) and multi-agent reinforcement learning (MARL) baselines. Analysis of interaction trajectories further reveals reduced reasoning redundancy and differentiated collaboration patterns characterized by questioning, complementing, and revising prior reasoning. Together, these results suggest that collaborative reasoning across heterogeneous models can improve task performance while enabling complementary contributions and the reassessment of prior reasoning across distinct model parameterizations.

Our key contributions are summarized in three points:

\begingroup
\setlength{\leftmargini}{1em}
\begin{itemize}
\setlength{\itemsep}{0pt}
\setlength{\parsep}{0pt}

\vspace{-0.7em}

\item We propose a trainable paradigm for collaborative reasoning and learning across heterogeneous models, enabling independent reassessment across distinct model parameterizations and progressive learning of effective cross-model collaboration from interaction trajectories.

\item Collaborative learning comprises model slicing and pyramid supervision. Model slicing constructs heterogeneous models under a comparable budget and enables reasoning-state isolation through task-state handoffs. Pyramid supervision combines interaction initialization with fine-grained supervision from the original large model over reasoning and interaction decisions.

\item We validate collaborative learning across four task categories. The trained heterogeneous models achieve comparable or better performance than the original large model, established MAS baselines, and MARL methods, while demonstrating higher inference efficiency than MAS and MARL baselines and exhibiting emergent task-adaptive cross-model collaboration patterns.

\end{itemize}
\endgroup
\footnotetext{Short linear, long linear, short cyclic, and long cyclic trajectories are defined in Section~\ref{sec:collaborative_pattern}.}

\vspace{-0.5em}
\section{Method}

\begin{figure}[!t]
\centering
\includegraphics[width=\linewidth]{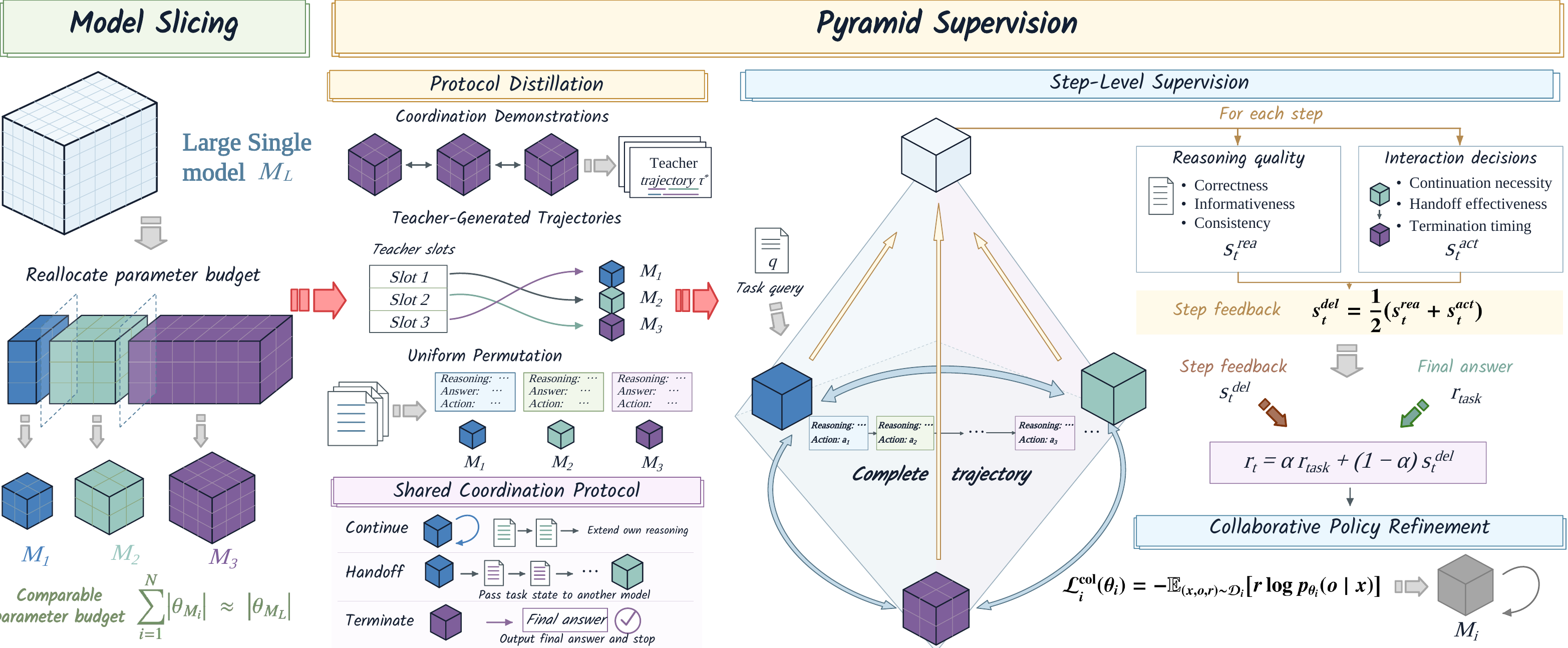}
\vspace{-1.25em}
\caption{\textbf{Overview of collaborative learning.} (a) Model slicing constructs smaller, heterogeneous models under a comparable total parameter budget. (b) Pyramid supervision initializes collaboration through protocol distillation and refines it through step-level supervision. The original large model evaluates each step’s reasoning and interaction decisions within the complete trajectory. Step-level scores and final task rewards jointly guide collaborative policy refinement.}
\label{fig:framework}
\end{figure}

Collaborative reasoning characterizes an iterative reasoning process among heterogeneous models, where explicit task-state handoffs under a unified interaction protocol allow prior reasoning to be continually reassessed by models with different parameters ($\triangleright$ \Cref{sec:collaborative_reasoning}). To realize this process, collaborative learning constructs a heterogeneous model set through model slicing and learns effective cross-model collaboration through pyramid supervision ($\triangleright$ \Cref{sec:collaborative_learning}).

\subsection{Collaborative Reasoning}

\label{sec:collaborative_reasoning}

Collaborative reasoning refers to a process in which multiple models jointly construct a solution through successive contributions~\citep{yin2023exchange,chen2024reconcile}. We define a unified interaction protocol that specifies how heterogeneous models generate reasoning and retain, transfer, or terminate control. For a task input $q$, each interaction step $t$ is represented by

\vspace{-1em}

\begin{equation}
z_t = (i_t, x_t, y_t, a_t)
\end{equation}
where $i_t$ denotes the active model $M_{i_t}$, $x_t$ contains the task input and preceding interaction history, $y_t$ is the current reasoning, and $a_t$ is the interaction action. The action space is defined as
\begin{equation}
a_t \in \{\texttt{continue},\texttt{terminate}\}
\cup \{\texttt{handoff}(j) \mid j \in \{1,\ldots,N\},\ j \neq i_t\}
\end{equation}
where \texttt{continue} keeps control with the model, \texttt{handoff(j)} transfers the task state and control to $M_j$, and \texttt{terminate} ends collaborative reasoning. A complete trajectory is represented as
\begin{equation}
\tau = (z_1,z_2,\ldots,z_T)
\end{equation}
where $T$ denotes the total number of interaction steps recorded in the complete trajectory.

Through \texttt{handoff}, models use distinct parameters to extend, question, or revise prior conclusions, enabling cross-model reassessment across the evolving reasoning process. The resulting trajectories record reasoning and interaction decisions for subsequent collaborative learning.

\subsection{Collaborative Learning}
\label{sec:collaborative_learning}

Collaborative learning encompasses both the construction of heterogeneous collaborators and the learning of cross-model collaboration, with the pipeline illustrated in Figure~\ref{fig:framework}. Model slicing constructs heterogeneous collaborators under a comparable parameter budget, providing the structural basis for cross-model reasoning. Pyramid supervision guides collaborative learning through protocol distillation and step-level supervision, progressively refining models' reasoning and interaction decisions while integrating task-level rewards throughout the collaborative reasoning process.

\subsubsection{Model Slicing}
\label{sec:model_slicing}

Model slicing replaces a large model with smaller, independently parameterized models under a comparable parameter budget, each operating as a standalone model and interacting through explicit task-state handoffs. Let $M_L$ denote the large model with parameters $\theta_L$. We have
\vspace{-0.3em}
\begin{equation}
\mathcal{M} = \{M_1, \ldots, M_N\},
\qquad \sum_{i=1}^{N} |\theta_i| \approx |\theta_L|
\end{equation}
where $\theta_i$ denotes the parameters of $M_i$, and $|\theta|$ denotes the parameter count.

These models differ in size and parameters and are not obtained by directly partitioning the weights of $M_L$. Together with task-state handoffs, this heterogeneous model set enables model-specific \textit{reasoning-state isolation} across distinct model parameterizations during collaboration.

\subsubsection{Pyramid Supervision}
\label{sec:Pyramid_Supervision}

We introduce \textit{pyramid supervision}, which refines collaboration through supervision across model scales. Protocol distillation uses the largest model slice to initialize cross-model interaction, while step-level supervision uses the original large model to evaluate collaboration trajectories.

\noindent\textbf{Protocol Distillation.}\quad
Although the interaction protocol specifies how models communicate and transfer control, heterogeneous models do not naturally learn to use it effectively as reasoning evolves. In particular, they must learn when to retain control, hand off the task state, or terminate based on the evolving reasoning context~\citep{zeng2024agenttuning}. We therefore use protocol distillation to initialize cross-model interaction from teacher-generated trajectories~\citep{chen2024magdi}.

We select the largest model in $\mathcal{M}$ by parameter count as the teacher $M^\star$ and instantiate $N$ homogeneous copies to form the teacher system:
\begin{equation}
    \mathcal{M}_{\mathrm{teacher}}
    =
    \{M_1^\star,\ldots,M_N^\star\},
    \qquad
    M_j^\star \equiv M^\star,\quad j=1,\ldots,N
\end{equation}

Following the same interaction protocol as the target models, the teacher system generates multi-step trajectories for distillation inputs
$\mathcal{Q}_{\mathrm{distill}}=\{q^{(m)}\}_{m=1}^{K}$:
\begin{equation}
    \tau^{(m)}
    \sim
    P_{\mathcal{M}_{\mathrm{teacher}}}
    \left(\tau \mid q^{(m)}\right),
    \qquad
    m=1,\ldots,K
\end{equation}
We retain qualified trajectories as demonstrations for subsequent distillation.

To avoid a fixed correspondence between teacher slots and target models, we sample a uniform permutation for each trajectory:
\begin{equation}
    \pi^{(m)}
    \sim
    \mathrm{Uniform}(S_N),
    \qquad
    \widetilde{i}_t^{(m)}
    =
    \pi^{(m)}
    \left(i_t^{(m)}\right),
    \quad
    t=1,\ldots,T_m
\end{equation}
where $S_N$ denotes the set of permutations over $\{1,\ldots,N\}$. Within each trajectory, the same permutation is applied consistently to all occurrences associated with each teacher slot, including model indices, handoff targets, and corresponding references in the interaction history. This operation changes only the correspondence between teacher slots and target models across trajectories, while preserving the reasoning content, step order, and handoff structure of the original trajectory.

Let $\widetilde{x}_t^{(m)}$ and $\widetilde{o}_t^{(m)}$ denote the reassigned context and response. Supervision covers the complete response, including reasoning and interaction decisions~\citep{chen2023fireact}. For each model $M_i$, the reassigned context--response pairs satisfying $\widetilde{i}_t^{(m)}=i$ form its training set $\mathcal{D}_i^{\mathrm{pro}}$. Following sequence-level knowledge distillation~\citep{kim2016sequence}, each model minimizes
\vspace{0.5em}
\begin{equation}
    \mathcal{L}_i^{\mathrm{pro}}
    =
    -\mathbb{E}_{(x,o)\sim\mathcal{D}_i^{\mathrm{pro}}}
    \left[
        \log p_{\theta_i}(o\mid x)
    \right]
\end{equation}

This objective initializes cross-model interaction capabilities in the heterogeneous models, providing a stable starting point for trajectory-level refinement of reasoning and interaction decisions.

\noindent\textbf{Step-Level Supervision.}\quad
After protocol distillation, the heterogeneous models perform collaborative reasoning under the shared interaction protocol, generating trajectories for subsequent optimization. For a given training task $q$, a collaborative trajectory is sampled as
\begin{equation}
    \tau
    =
    (z_1,\ldots,z_T)
    \sim
    P_{\mathcal{M}}(\tau\mid q)
\end{equation}
where $P_{\mathcal{M}}$ is induced by the current models and the shared interaction protocol.

Learning effective collaboration from these trajectories requires identifying the contribution of each reasoning and interaction decision to the overall process~\citep{foerster2018coma}. The value of an individual step depends not only on its local correctness but also on how it influences subsequent collaboration. We therefore retain the original large model $M_L$ as an evaluator $\mathcal{J}$, which assesses each step in the context of the complete trajectory. Given the task $q$ and trajectory $\tau$, the evaluator produces two complementary scores:
\begin{equation}
    \left\{
    \left(
    s_t^{\mathrm{rea}},
    s_t^{\mathrm{act}}
    \right)
    \right\}_{t=1}^{T}
    =
    \mathcal{J}(q,\tau),
    \qquad
    s_t^{\mathrm{rea}},s_t^{\mathrm{act}}\in[-1,1]
\end{equation}

The reasoning score $s_t^{\mathrm{rea}}$ evaluates the model's reasoning contribution in terms of correctness, informativeness, and relevance to the task state and interaction history. The interaction score $s_t^{\mathrm{act}}$ evaluates whether its decision to continue, hand off the task state, or terminate appropriately advances the collaborative process. Separating these dimensions avoids conflating reasoning quality with interaction quality. We average the two scores to obtain a step-level score:
\begin{equation}
    s_t^{\mathrm{del}}
    =
    \frac{1}{2}
    \left(
    s_t^{\mathrm{rea}}
    +
    s_t^{\mathrm{act}}
    \right)
\end{equation}
\vspace{-1em}

\noindent\textit{Reward integration.}\quad
Step-level scores provide fine-grained supervision for intermediate reasoning and interaction decisions, but local decision quality does not necessarily translate into successful task completion. To provide decision-specific feedback while also accounting for the overall task objective, we integrate each step-level score with the final task reward. Let $m(q,\hat{y})\in[0,1]$ denote the task-specific evaluation metric for the final prediction $\hat{y}$. We map it to the same $[-1,1]$ range as the step-level score and define
\begin{equation}
    r_{\mathrm{task}}
    =
    2m(q,\hat{y})-1,
    \qquad
    r_t
    =
    \alpha r_{\mathrm{task}}
    +
    (1-\alpha)s_t^{\mathrm{del}}
    \label{eq:step_reward}
\end{equation}
where $\alpha\in[0,1]$ controls the trade-off between the task-level reward and the step-level score. The resulting reward distinguishes the contributions of intermediate steps across the complete collaborative trajectory while aligning their optimization with the final task objective.

\noindent\textit{Collaborative Policy Refinement.}\quad
Using the combined rewards, we extract the steps at which each model was active from the collaborative trajectories and construct model-specific training data. For model $M_i$, the corresponding training examples are collected as

\begin{equation}
\mathcal{D}_i
=
\left\{
(x_t^{(m)},o_t^{(m)},r_t^{(m)})
\mid
i_t^{(m)}=i
\right\}  
\end{equation}
where $o_t^{(m)}$ denotes the complete response at step $t$ of trajectory $m$, serialized under the shared interaction protocol and including its reasoning, answer, and interaction fields. Following the score-function policy-gradient form, we optimize each model with the reward-weighted objective
\begin{equation}
    \mathcal{L}_i^{\mathrm{col}}(\theta_i)
    =
    -\mathbb{E}_{(x,o,r)\sim\mathcal{D}_i}
    \left[
        r\log p_{\theta_i}(o\mid x)
    \right]
    \label{eq:collaborative_objective}
\end{equation}

This objective increases the likelihood of responses assigned positive rewards and decreases the likelihood of those assigned negative rewards. Because each reward combines step-level evaluation conditioned on the complete trajectory with the final task outcome, each model can adapt its policy according to how its own decisions contribute to the collaborative result, allowing differentiated collaboration patterns to progressively emerge through training~\citep{cui2026prime}.

\section{Experiments}
\label{sec:experiments}

\subsection{Experimental Setup}
\label{sec:experimental_setup}

\noindent\textbf{Datasets.}\quad We use five benchmarks across four task categories: multi-hop question answering (MuSiQue~\citep{trivedi2022musique}), mathematical reasoning (GSM-Hard~\citep{gao2023pal} and MATH~\citep{hendrycks2021math}), code generation (an eight-language MultiPL-E subset~\citep{cassano2023multiple})\footnote{Following the language selection in the Qwen3 technical report~\citep{yang2025qwen3}, we restrict code-generation evaluation to Python, C++, Java, PHP, TypeScript, C\#, Bash, and JavaScript.}, and open-ended constrained generation (Conifer~\citep{sun2024conifer}).

\noindent\textbf{Baselines.}\quad
(1) Single-agent baselines: Qwen3-14B~\citep{yang2025qwen3} and its self-evaluated RL variant (SE-RL)\footnote{SE-RL uses Qwen3-14B self-feedback for comparison with cross-model feedback.}.
(2) A homogeneous MAS of three Qwen3-8B models (24B total) without collaborative training, serving as a higher-capacity multi-model baseline.
(3) A homogeneous collaborative-learning system (Homo. \lmttbf{CL}) of three Qwen3-4B models (12B total), trained with the same procedure as our heterogeneous system to assess the effect of model heterogeneity in collaborative reasoning.
(4) Established multi-agent frameworks: MAD~\citep{du2024debate}, AgentVerse~\citep{chen2024agentverse}, AFlow~\citep{zhang2025aflow}, and GPTSwarm~\citep{zhuge2024gptswarm}.
(5) MARL methods: MAGRPO~\citep{liu2026collaboration}, MAPoRL~\citep{park2025maporl}, and AT-GRPO~\citep{zhao2026strongermas}.

\noindent\textbf{Metrics.}\quad We report accuracy and F1 on MuSiQue~\citep{trivedi2022musique}, GSM-Hard~\citep{gao2023pal}, and MATH~\citep{hendrycks2021math}. We use weighted pass@1 and macro-language pass@1 for MultiPL-E~\citep{cassano2023multiple}, and Coverage and Explicit for Conifer~\citep{sun2024conifer}. Beyond task performance, we analyze handoffs, anomaly rates, reasoning redundancy, and model-wise behavioral distributions to characterize training-induced changes in collaboration patterns.

\noindent\textbf{Implementation Details.}\quad
Model slicing replaces Qwen3-14B ($M_L$)~\citep{yang2025qwen3} with Qwen3-1.7B, Qwen3-4B, and Qwen3-8B (13.7B total). The three collaborating models share the same prompt and action space. Three Qwen3-8B instances generate protocol distillation demonstrations, while Qwen3-14B evaluates reasoning and interaction decisions within complete trajectories for step-level supervision. Experiments run on $8\times$ NVIDIA A800-SXM4-80GB GPUs. Each model is trained with a rank-64 LoRA adapter~\citep{hu2022lora} while its base parameters remain frozen.

\vspace{-0.2em}

\subsection{Does Collaborative Learning Lead to Improved Performance?}
\label{sec:main_results}

\vspace{-1.5em}

\begin{table}[!htbp]
\centering
\vspace{0.4em}

\caption{\textbf{Performance comparison with baseline methods.} Scores are percentages (\%); higher is better. Acc. denotes accuracy; Wtd. and Mac. denote weighted and macro-language pass@1, respectively; Cov. and Expl. denote Coverage and Explicit, respectively. \lmttbf{CL.} denotes collaborative learning. Best and second-best results are shown in \textbf{bold} and \underline{underline}, respectively. $\textcolor[RGB]{241,96,52}{\uparrow}$ and $\textcolor[RGB]{1,179,185}{\downarrow}$ indicate scores above and below the corresponding baseline mean for each metric in the table. $\dagger$ indicates that the corresponding baseline differs significantly from \lmttbf{CL.} ($p \leq 0.05$).}
\label{tab:main_results}
\begingroup
\fontsize{12}{14}\selectfont
\setlength{\tabcolsep}{2pt}
\renewcommand{\arraystretch}{1.32}
\setlength{\arrayrulewidth}{0.4pt}
\newcommand{\ResultUp}[1]{\ensuremath{{}_{\textcolor[RGB]{241,96,52}{\scalebox{1.1}{$\scriptstyle\uparrow\!#1$}}}}}
\newcommand{\ResultDown}[1]{\ensuremath{{}_{\textcolor[RGB]{1,179,185}{\scalebox{1.1}{$\scriptstyle\downarrow\!#1$}}}}}
\newcommand{\MainResultsHeaderStrut}{\rule[-0.45em]{0pt}{1.85em}}
\vspace{4pt}
\resizebox{\linewidth}{!}{%
\begin{NiceTabular}{l!{\color{black}\vrule width 0.6pt}*{10}{l}}
\CodeBefore
  \rectanglecolor{CadetBlue!20}{1-1}{2-11}
  \rowcolor{gray!10}{4,6,8,10,12,14}
\Body
\arrayrulecolor{black}
\Xhline{1pt}
\Block[l]{2-1}{\textbf{Method}} & \multicolumn{2}{c}{\MainResultsHeaderStrut\textbf{MuSiQue}} & \multicolumn{2}{c}{\MainResultsHeaderStrut\textbf{GSM-Hard}} & \multicolumn{2}{c}{\MainResultsHeaderStrut\textbf{MATH}} & \multicolumn{2}{c}{\MainResultsHeaderStrut\textbf{MultiPL-E}} & \multicolumn{2}{c}{\MainResultsHeaderStrut\textbf{Conifer}} \\
& {\small\MainResultsHeaderStrut\textbf{Acc.}} & {\small\MainResultsHeaderStrut\textbf{F1}} & {\small\MainResultsHeaderStrut\textbf{Acc.}} & {\small\MainResultsHeaderStrut\textbf{F1}} & {\small\MainResultsHeaderStrut\textbf{Acc.}} & {\small\MainResultsHeaderStrut\textbf{F1}} & {\small\MainResultsHeaderStrut\textbf{Wtd.}} & {\small\MainResultsHeaderStrut\textbf{Mac.}} & {\small\MainResultsHeaderStrut\textbf{Cov.}} & {\small\MainResultsHeaderStrut\textbf{Expl.}} \\
\Xhline{1pt}
Qwen3-14B & 39.55$^{\dagger}$\ResultUp{0.87} & 49.70$^{\dagger}$\ResultUp{0.47} & 68.94\ResultUp{1.65} & \underline{80.69}$^{\dagger}$\ResultUp{3.19} & 74.60$^{\dagger}$\ResultUp{4.40} & 82.86$^{\dagger}$\ResultUp{3.25} & 55.18$^{\dagger}$\ResultUp{1.08} & 57.34$^{\dagger}$\ResultUp{2.59} & 88.00$^{\dagger}$\ResultUp{0.17} & 95.72\ResultUp{0.62} \\

SE-RL & 41.75\ResultUp{3.07} & 52.71$^{\dagger}$\ResultUp{3.48} & 68.18\ResultUp{0.89} & 79.67$^{\dagger}$\ResultUp{2.17} & \underline{76.00}\ResultUp{5.80} & 80.70\ResultUp{1.09} & \textbf{60.43}$^{\dagger}$\ResultUp{6.33} & \textbf{62.55}$^{\dagger}$\ResultUp{7.80} & 88.00\ResultUp{0.17} & 95.81\ResultUp{0.71} \\

\Xhline{0.2pt}

Homo. MAS & 26.40$^{\dagger}$\ResultDown{12.28} & 32.82\ResultDown{16.41} & \underline{71.21}\ResultUp{3.92} & 79.93$^{\dagger}$\ResultUp{2.43} & 68.20$^{\dagger}$\ResultDown{2.00} & 76.41$^{\dagger}$\ResultDown{3.20} & 56.36$^{\dagger}$\ResultUp{2.26} & 57.20\ResultUp{2.45} & 88.02$^{\dagger}$\ResultUp{0.19} & 94.27$^{\dagger}$\ResultDown{0.83} \\

\Xhline{0.2pt}

Homo. \lmttbf{CL.} & 36.00$^{\dagger}$\ResultDown{2.68} & 47.19$^{\dagger}$\ResultDown{2.04} & 60.61$^{\dagger}$\ResultDown{6.68} & 76.51\ResultDown{0.99} & 68.60$^{\dagger}$\ResultDown{1.60} & 77.95\ResultDown{1.66} & 53.33$^{\dagger}$\ResultDown{0.77} & 52.65$^{\dagger}$\ResultDown{2.10} & 88.15\ResultUp{0.32} & 95.22$^{\dagger}$\ResultUp{0.12} \\

\Xhline{0.2pt}

MAD & 39.51$^{\dagger}$\ResultUp{0.83} & 50.59$^{\dagger}$\ResultUp{1.36} & 66.67$^{\dagger}$\ResultDown{0.62} & 74.18\ResultDown{3.32} & 72.40$^{\dagger}$\ResultUp{2.20} & 81.32\ResultUp{1.71} & 55.03$^{\dagger}$\ResultUp{0.93} & 55.35$^{\dagger}$\ResultUp{0.60} & 88.55\ResultUp{0.72} & 95.84\ResultUp{0.74} \\

AgentVerse & \underline{41.99}\ResultUp{3.31} & 52.85$^{\dagger}$\ResultUp{3.62} & \textbf{71.97}\ResultUp{4.68} & 78.44$^{\dagger}$\ResultUp{0.94} & 68.20$^{\dagger}$\ResultDown{2.00} & 79.42\ResultDown{0.19} & 59.25\ResultUp{5.15} & \underline{59.83}$^{\dagger}$\ResultUp{5.08} & 87.74$^{\dagger}$\ResultDown{0.09} & 95.86$^{\dagger}$\ResultUp{0.76} \\

AFlow & 41.83$^{\dagger}$\ResultUp{3.15} & 50.36$^{\dagger}$\ResultUp{1.13} & 68.18\ResultUp{0.89} & 76.04\ResultDown{1.46} & 65.00$^{\dagger}$\ResultDown{5.20} & 76.53$^{\dagger}$\ResultDown{3.08} & 42.83\ResultDown{11.27} & 42.46$^{\dagger}$\ResultDown{12.29} & 87.92$^{\dagger}$\ResultUp{0.09} & \textbf{96.51}$^{\dagger}$\ResultUp{1.41} \\

GPTSwarm & 36.20$^{\dagger}$\ResultDown{2.48} & 47.37$^{\dagger}$\ResultDown{1.86} & 70.45\ResultUp{3.16} & 78.39$^{\dagger}$\ResultUp{0.89} & 69.20$^{\dagger}$\ResultDown{1.00} & 79.90\ResultUp{0.29} & 51.63$^{\dagger}$\ResultDown{2.47} & 52.43\ResultDown{2.32} & 88.17$^{\dagger}$\ResultUp{0.34} & \underline{95.96}$^{\dagger}$\ResultUp{0.86} \\

\arrayrulecolor{black}

\Xhline{0.2pt}

MAGRPO & 40.13$^{\dagger}$\ResultUp{1.45} & 51.74\ResultUp{2.51} & 65.91\ResultDown{1.38} & 76.44\ResultDown{1.06} & 65.00$^{\dagger}$\ResultDown{5.20} & 76.01$^{\dagger}$\ResultDown{3.60} & 49.93\ResultDown{4.17} & 50.48$^{\dagger}$\ResultDown{4.27} & \underline{88.58}$^{\dagger}$\ResultUp{0.75} & 95.82$^{\dagger}$\ResultUp{0.72} \\

MAPoRL & 40.59$^{\dagger}$\ResultUp{1.91} & 52.74$^{\dagger}$\ResultUp{3.51} & 63.64$^{\dagger}$\ResultDown{3.65} & 75.34$^{\dagger}$\ResultDown{2.16} & 70.80$^{\dagger}$\ResultUp{0.60} & 80.87$^{\dagger}$\ResultUp{1.26} & 53.25$^{\dagger}$\ResultDown{0.85} & 53.27$^{\dagger}$\ResultDown{1.48} & 86.37$^{\dagger}$\ResultDown{1.46} & 89.19$^{\dagger}$\ResultDown{5.91} \\

AT-GRPO & 41.54\ResultUp{2.86} & \textbf{53.51}$^{\dagger}$\ResultUp{4.28} & 64.39$^{\dagger}$\ResultDown{2.90} & 76.86\ResultDown{0.64} & 74.20$^{\dagger}$\ResultUp{4.00} & \underline{83.70}$^{\dagger}$\ResultUp{4.09} & 57.91$^{\dagger}$\ResultUp{3.81} & 58.74$^{\dagger}$\ResultUp{3.99} & 86.60$^{\dagger}$\ResultDown{1.23} & 95.85$^{\dagger}$\ResultUp{0.75} \\

\Xhline{0.2pt}

\lmttbf{CL.} & \textbf{42.37}\ResultUp{3.69} & \underline{53.43}\ResultUp{4.20} & \textbf{71.97}\ResultUp{4.68} & \textbf{81.15}\ResultUp{3.65} & \textbf{78.40}\ResultUp{8.20} & \textbf{85.25}\ResultUp{5.64} & \underline{59.39}\ResultUp{5.29} & 59.63\ResultUp{4.88} & \textbf{88.90}\ResultUp{1.07} & 95.90\ResultUp{0.80} \\
\Xhline{1pt}
\end{NiceTabular}}
\endgroup
\end{table}
\FloatBarrier

\vspace{-0.5em}

As shown in Table~\ref{tab:main_results}, \textbf{collaborative learning achieves the highest average performance across five benchmarks, with best or joint-best results on most metrics}. With a total parameter budget of 13.7B, it outperforms the original Qwen3-14B on all ten metrics, with an average absolute gain of 2.38\%, and surpasses SE-RL on eight. It also outperforms the 24B homogeneous MAS without collaborative training on all ten metrics, despite using fewer parameters. Compared with Homo. CL, which trains homogeneous models using the same procedure, the heterogeneous collaborative system achieves higher scores across all ten metrics, with an average absolute gain of 6.02\%. Its average absolute gains over established MAS and MARL baselines range from 2.08\% to 6.87\% across the ten metrics, demonstrating competitive performance against both classes.

Trajectory analysis reveals task-specific characteristics of collaboration and how intermediate behaviors translate into final outcomes. On MuSiQue, subsequent interactions correct initial errors, while on MATH, the trained heterogeneous models better preserve correct intermediate answers than MAPoRL. On MuSiQue and GSM-Hard, differences in partial-match scores on questions missed by both the collaborative system and the baseline help explain why accuracy and F1 rankings do not fully align. Code-generation performance varies across task subsets: the collaborative system solves more MBPP tasks but fewer HumanEval tasks than SE-RL. On most tasks solved only by SE-RL, it produces no passing intermediate program, suggesting that this gap primarily reflects failure to generate correct solutions rather than the loss of correct candidates during subsequent interactions. On Conifer, the collaborative system achieves the highest Coverage and almost always maintains compliance in the final output when an intermediate answer already satisfies all constraints.

\vspace{-0.2em}

\subsection{How Much Does Each Mechanism Contribute to Performance?}
\label{sec:ablation}

\begin{figure}[!t]
\centering
\includegraphics[width=\linewidth]{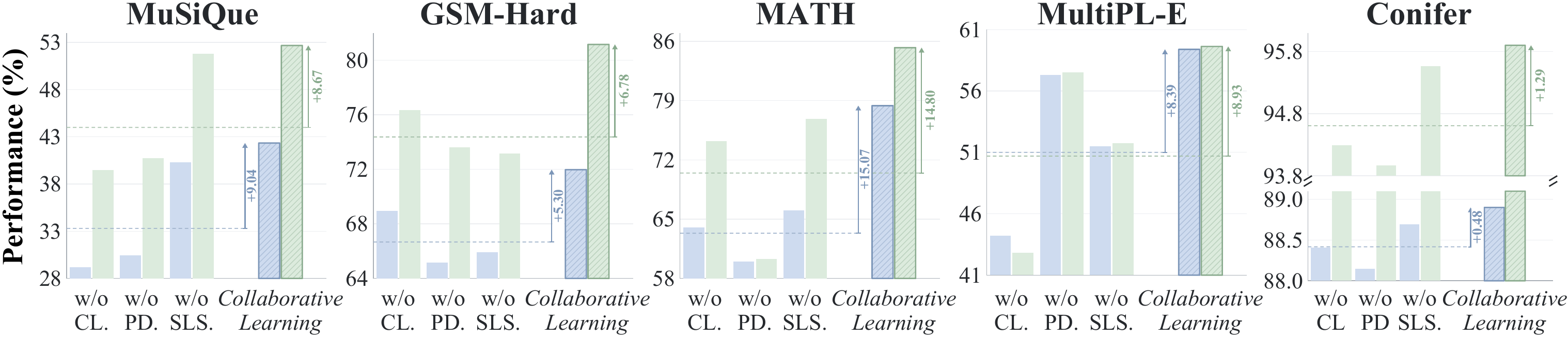}
\vspace{-1.5em}
\caption{\textbf{Ablation study of the mechanism across five benchmarks.} We compare a zero-shot variant that receives no training (w/o CL.), a variant without Protocol Distillation (w/o PD.), and a variant without the step-level score (w/o SLS.). Arrows indicate the gain of \textit{collaborative learning} over the arithmetic mean of the available variants. For each benchmark, \raisebox{0.15ex}{\textcolor[RGB]{206,219,239}{\rule{1.3em}{0.8em}}} denotes Acc., Wtd., and Cov., while \raisebox{0.15ex}{\textcolor[RGB]{220,235,221}{\rule{1.3em}{0.8em}}} denotes the corresponding second metric (F1, Mac., or Expl.).}
\label{fig:ablation_results}
\end{figure}

As shown in Figure~\ref{fig:ablation_results}, we compare three ablation variants: without collaborative training (w/o CL.), without protocol distillation (w/o PD.), and without the step-level score (w/o SLS.). Averaged over the ten reported metrics, these variants incur absolute score drops of 9.37, 8.85, and 5.41 points, respectively, and each underperforms the full method on every metric. These results indicate that \textbf{heterogeneity alone is insufficient to achieve the full performance gains, and that both protocol distillation and step-level scoring contribute meaningfully to learned collaboration}.

\FloatBarrier

Beyond aggregate performance, trajectory-level behavior provides a more direct view of how the two training components shape collaboration throughout collaborative reasoning. For example, without protocol distillation, first-answer EM on MuSiQue increases from 21.02\% to 25.24\%, yet the number of initially incorrect answers corrected by the final response falls from 594 to 171. Meanwhile, the proportion of Conifer trajectories that initially satisfy all constraints but fail at termination increases, and protocol-following failures on MATH also become more frequent. These observations suggest that protocol distillation primarily supports effective downstream interaction rather than simply improving the quality of the initial response. Similarly, removing the step-level score increases the average MuSiQue trajectory length from 3.4 to 3.8 steps while reducing the number of corrected answers by 18.5\%, indicating that step-level feedback helps models use interaction steps more productively and achieve more corrections with fewer interactions on this task.

\vspace{-0.4em}
\subsection{How Efficient and Effective Is Collaborative Reasoning?}
\label{sec:efficiency}

\begin{figure}[!htbp]
\centering
\includegraphics[width=\linewidth]{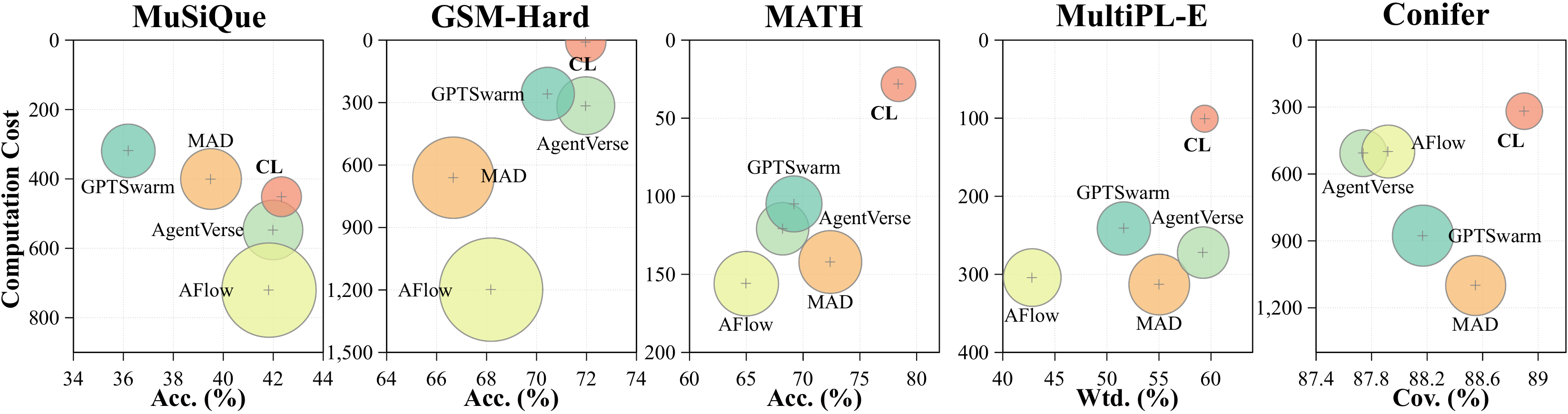}
\vspace{-1.3em}
\caption[Computation cost versus task performance.]
{\textbf{Computation cost versus task performance.}
The center of each circle indicates task performance and parameter-weighted output tokens
for the collaborative system and four multi-agent baselines across five benchmarks;
its area is proportional to the number of model invocations.}
\label{fig:efficiency}
\end{figure}

Alongside improvements in task performance, collaborative learning leads to more efficient inference. As shown in Figure~\ref{fig:efficiency}, the collaborative system requires fewer model invocations than MAD, AgentVerse, AFlow, and GPTSwarm across all five benchmarks, with reductions ranging from 39.1\% to 84.8\%. Specifically, it requires an average of 4.0, 4.1, 1.8, 3.0, and 3.3 invocations per question on MuSiQue, GSM-Hard, MultiPL-E, MATH, and Conifer, respectively. This efficiency advantage also extends to the evaluated MARL baselines, with invocation reductions ranging from 46.7\% to 86.0\%. In terms of parameter-weighted output tokens, \textbf{the collaborative system achieves the lowest count among the compared methods on four of five benchmarks}. Compared with the MARL baselines, it further reduces parameter-weighted output tokens by 8.2\%--89.4\%.

We further compare reasoning-overlap rates before and after collaborative learning to examine redundancy across successive reasoning steps. On MuSiQue, the overlap rates of the three collaborating models decrease from 48.3\%, 24.1\%, and 55.2\% to 0.2\%, 0.3\%, and 0.5\%, respectively. On MultiPL-E, they decrease from 34.8\%, 2.3\%, and 17.6\% to 7.4\%, 1.8\%, and 10.7\%. These results suggest that \textbf{collaborative learning substantially reduces repetitive reasoning across model interactions, yielding less redundant subsequent contributions throughout the process}.

To further assess collaborative effectiveness, we measure correct-answer retention, defined as the proportion of trajectories whose final answer remains correct among those containing a correct intermediate candidate. On MuSiQue, GSM-Hard, MultiPL-E, and MATH, the trained heterogeneous models achieve 80.42\%, 97.94\%, 98.26\%, and 98.00\% retention, respectively, ranking first on the latter two tasks, second on GSM-Hard, and third on MuSiQue among comparison methods with recoverable intermediate candidates\footnote{Conifer is excluded because its open-ended outputs lack an intermediate-candidate correctness criterion.}. These results show that \textbf{the trained models generally preserve available correct intermediate answers throughout subsequent collaborative interactions}.

\FloatBarrier
\vspace{-1em}

\subsection{What Collaboration Patterns Emerge from Collaborative Learning?}
\label{sec:collaborative_pattern}

\begin{figure}[!htbp]
\centering
\includegraphics[width=\linewidth]{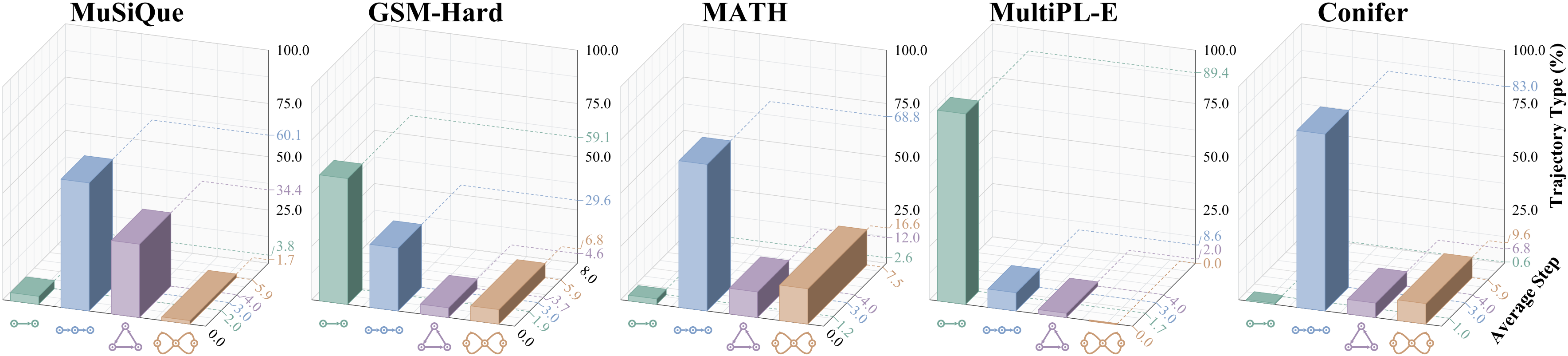}
\vspace{-1.6em}
\caption{\textbf{Post-training interaction topologies across tasks.}
\protect\raisebox{-0.15em}{\protect\includegraphics[height=1.15em]{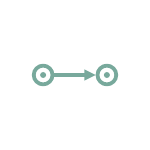}},
\protect\raisebox{-0.15em}{\protect\includegraphics[height=1.15em]{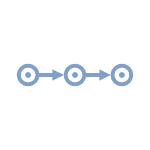}},
\protect\raisebox{-0.15em}{\protect\includegraphics[height=1.15em]{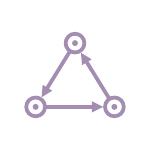}},
and
\protect\raisebox{-0.15em}{\protect\includegraphics[height=1.15em]{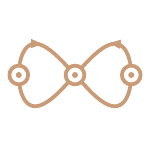}}
denote short linear, long linear, short cyclic, and long cyclic trajectories, respectively. Bar height indicates the proportion of each trajectory type, while depth indicates the average number of interaction steps.}
\label{fig:interaction_topology}
\end{figure}
\FloatBarrier

\vspace{-0.5em}

To examine how collaborative learning reorganizes model interactions across tasks, we compare trajectory topologies and interaction lengths before and after training. We categorize trajectories as short linear, long linear, short cyclic, or long cyclic. Linear trajectories involve each model at most once, with at most two interaction steps considered short; cyclic trajectories revisit a previously participating model, with at most four steps considered short. On MuSiQue, the proportion of long linear trajectories increases from 9.18\% to 60.12\%, showing a shift toward longer linear interactions. In contrast, this proportion decreases from 10.95\% to 8.58\% on MultiPL-E. Short cyclic trajectories evolve differently across tasks: their average interaction length decreases from 4.00 to 3.67 steps on GSM-Hard, but increases from 3.33 to 4.00 on MATH. These results show that collaborative learning does not uniformly increase interaction complexity or length, but instead induces task-dependent collaboration topologies rather than converging to a single universal interaction pattern.

\textbf{Task-dependent functional specialization also emerges across models during collaborative learning.} On MuSiQue, Qwen3-4B's answer-revision rate increases from 13\% to 69\%, its handoff rate from 48\% to 96\%, and its termination rate decreases from 50\% to 4\%, indicating a shift toward intermediate revision and handoff. Meanwhile, Qwen3-8B increasingly provides the final answer, with its share rising from 13.0\% to 58.4\%. This division of labor changes across tasks: on GSM-Hard, Qwen3-4B instead becomes the more frequent final-answer provider, with its share increasing from 34.09\% to 64.39\%. Thus, the same model can assume different functions across tasks, revealing task-dependent divisions of labor rather than a fixed collaboration structure.

\vspace{-0.4em}
\section{Related Work}
\label{sec:related_work}

\noindent\textbf{Reasoning in LLMs.}\quad Existing work improves LLM reasoning by eliciting existing capabilities or strengthening them through training~\citep{zhao2026selfdistilled,li2026rise,liu2026spade}. Inference-time methods use chain-of-thought prompting~\citep{wei2022chain}, systematic exploration~\citep{hariri2026regimes,li2026tailsearch}, evaluation~\citep{wu2026stv}, and revision~\citep{yang2026retree}; training-based methods learn from bootstrapped demonstrations~\citep{zelikman2022star,dong2026bird}, distill teacher reasoning~\citep{yang2025supercorrect,liu2026sparse,fu2026onetraining}, or use outcome- and process-level supervision. Yet these methods remain bounded by a single model's knowledge and reasoning biases, leaving blind spots unaddressed and allowing intermediate errors to persist through subsequent reasoning~\citep{xie2026breadth}.

\noindent\textbf{From Single-Model Reasoning to Multi-Model Composition.}\quad These limitations motivate moving beyond single-model reasoning toward multi-model composition, which exploits complementary knowledge and task-specific strengths across models~\citep{yang2026corl,zhang2026hacrl,huot2026meaningful}. Existing approaches select models suited to an input~\citep{ong2025routellm,cao2026scope,feng2026llmrouter,fisch2026pandora}, distribute tasks or subtasks~\citep{agashe2025agents2,chen2026mhgpo}, organize sequential model invocations~\citep{dekoninck2025routing,li2026progrouter}, determine when additional models are needed~\citep{rajib2026gatedmemory,qin2026atlas}, and aggregate candidate outputs~\citep{zhang2026agentrouter}. However, system-level composition does not necessarily constitute collaborative reasoning in which models jointly develop, evaluate, and revise a shared solution.

\noindent\textbf{Multi-Model Collaborative Reasoning.}\quad Collaborative reasoning directly addresses how models jointly develop solutions through interaction. Existing approaches range from hand-designed roles and protocols to optimized workflows~\citep{hu2025adas,zhou2026mass,lee2026mega,koh2026graft} and learned collaborative policies~\citep{xue2026comas,zhang2026metaagentx}. Representative systems, however, often retain an explicitly prescribed division of labor~\citep{zhao2026strongermas,feng2026drmas,ding2026apex}. In contrast, we train heterogeneous small models to collaborate without predefined roles under a total parameter budget comparable to that of a single large model.

\vspace{-0.5em}

\section{Conclusion}

\looseness=-1

We introduced a collaborative learning framework for reasoning across heterogeneous models. Model slicing replaces a large model with smaller heterogeneous models under a comparable parameter budget, enabling reasoning-state isolation through task-state handoffs. Pyramid supervision enables cross-model collaboration by combining protocol distillation for interaction initialization with step-level supervision over model-generated trajectories. Across four task categories, the trained heterogeneous models match or outperform the large model and MAS and MARL baselines, while requiring fewer model invocations than multi-agent frameworks. Analysis reveals reduced reasoning redundancy, stronger correct-answer retention, and task-dependent interaction topologies and functional specialization. These results suggest that collaborative learning can transform model heterogeneity into complementary reasoning capabilities across distinct model parameterizations.

Our study covers three model sizes and five benchmarks, leaving generalization to larger model pools, longer interaction horizons, and unseen task distributions unexplored. Pyramid supervision depends on inspector and evaluator quality, whose biases may affect step-level feedback. Comparable parameter counts do not imply equal computational cost, as collaborative inference adds latency and memory overhead beyond our parameter-weighted output-token metric. Future work will study generalization and robust supervision through multiple inspectors or uncertainty-aware aggregation.

\begingroup
\raggedright
\bibliography{iclr2027_conference}
\bibliographystyle{iclr2027_conference}
\endgroup

\end{document}